\documentclass[review,12pt]{elsarticle}

\usepackage{graphicx}
\usepackage{epstopdf}
\usepackage{amsmath}
\usepackage{picins}

\usepackage{amssymb}
\journal{Neurocomputing Letters}

\begin{document}

\begin{frontmatter}


\title{Storing non-uniformly distributed messages in networks of neural cliques\tnoteref{t1}}
\tnotetext[t1]{This work was supported in part by the European Research Council (ERC-AdG2011 290901 NEUCOD)}
\author[cea,tb]{Bartosz~Boguslawski\corref{cor1}}
\ead{bartosz.boguslawski@cea.fr}
\author[tb]{Vincent~Gripon}
\author[tb]{Fabrice~Seguin}
\author[cea]{Fr\'{e}d\'{e}ric~Heitzmann}
\cortext[cor1]{Corresponding author. Tel.: +33 4 38 78 01 06}

\address[cea]{CEA Leti - MINATEC, 17 rue des Martyrs, 38054 GRENOBLE CEDEX 9, France}
\address[tb]{TELECOM Bretagne - Electronics Department, CNRS Lab-STICC UMR 3192, Techn\^{o}pole Brest Iroise-CS 83818, 29238 BREST CEDEX 3, France}

\author{}

\address{}

\begin{abstract}
Associative memories are data structures that allow retrieval of stored messages from part of their content. They thus behave similarly to human brain that is capable for instance of retrieving the end of a song given its beginning. Among different families of associative memories, sparse ones are known to provide the best efficiency (ratio of the number of bits stored to that of bits used). Nevertheless, it is well known that non-uniformity of the stored messages can lead to dramatic decrease in performance. We introduce several strategies to allow efficient storage of non-uniform messages in recently introduced sparse associative memories. We analyse and discuss the methods introduced. We also present a practical application example.
\end{abstract}


\end{frontmatter}

% \linenumbers

%\section{}
%\label{}
\section{Introduction}

Traditional indexed memories allow retrieving data by using a unique address pointing to the searched data. The principle of associative memories is different. Data retrieval is accomplished by presenting a part (possibly small) of searched data. Thanks to the partial input, the rest of the information is recalled and consequently no address is needed. Associative memories are widely used in practical applications, for instance databases \cite{Lin:1976}, intrusion detection \cite{Papadogiannakis:2010}, processing units' caches \cite{Jouppi:1990} or routers \cite{Huang:2001}.

In domain of associative memories several characteristics can be defined in order to describe the performance of this specific approach. Here, four parameters are taken into account \cite{GriRab20139}. The first is the memory efficiency which is defined as the best ratio between the maximum amount of stored data in bits and the amount of information used for a given error rate. The second one is the retrieval error rate. The two last ones are the computational complexity which leads to certain energy efficiency, and the flexibility understood as the capability to react to non-exact input data.

In the category of associative memories one can distinguish two main solutions, namely content-addressable memory (CAM) \cite{pagiamtzis-jssc:2006} and neuroinspired memories. A CAM compares the input search word against the stored data, and returns a list of one or more addresses where the matching data word is stored. CAMs combine memory efficiency with zero error rate. However, the number of comparisons between the input search word and stored data results in high complexity and energy consumption \cite{JarollahiGripon:2013}. Moreover, CAMs are not flexible since the input partial erasures prevent correct address association. Ternary CAM (TCAM) expands CAM functionality offering more flexibility. However, this comes at an additional cost as the memory cells must encode three possible states instead of the two in case of binary CAM. Consequently, the cost of parallel search within such a memory becomes even greater \cite{Agrawal06tcam}.

Neuroinspired associative memories combine lower complexity with higher flexibility. In this category Hopfield network \cite{HopfTank:82} is the most prominent model. Nevertheless, when the size of this network is increased the efficiency tends to zero for a given error rate. Sparse networks proposed by Willshaw \cite{willshaw:nonholographic} are known to provide the best efficiency \cite{Palm:2013}.

At present, associative memories are exploited mainly on the hardware level. Nevertheless, the lack of a solution combining the four above-mentioned characteristics inhibits the use of associative memories on the algorithmic, software level.

Recently Gripon and Berrou proposed a new model \cite{GriBer20117} that can actually be seen as a particular Willshaw network with cluster structure. This modification allows for efficient retrieval algorithm without diminishing performance. It offers a good trade off providing high efficiency and low error rate combined with low complexity and increased flexibility. This model is able to store a large number of messages and recall them, even when a significant part of the input is erased. The evaluation of the network working as a data structure or an associative memory proved a huge gain in performance compared to Hopfield network \cite{HopfTank:82} and Boltzmann machine \cite{AcklHint:85} (when using comparable material). Moreover, the network is able to retrieve messages with erasures on random positions which gives it an advantage over CAMs.

The network as presented in \cite{GriBer20117} is analyzed only for uniform i.i.d. (independent identically distributed) messages. By the network construction this means that the number of connections going out from each node is uniformly distributed across the whole network. It is well known that non-uniformity of messages to store can lead to dramatic decrease in performance \cite{Knoblauch:2010}. In this work the situation where non-uniform data is stored is analyzed and its influence on the network performance is explained. Further, several techniques solving the problem of storing such data are proposed.

The paper is organized as follows. Section II outlines the theory of the network. The problem of storing non-uniform data is explained in Section III. Section IV proposes several new strategies to store such data. In Section V performance of the proposed stategies is evaluated. Then, an application example is given.

\section{Sparse neural networks with large learning diversity}

\begin{figure}
\centering
\includegraphics[width=70mm]{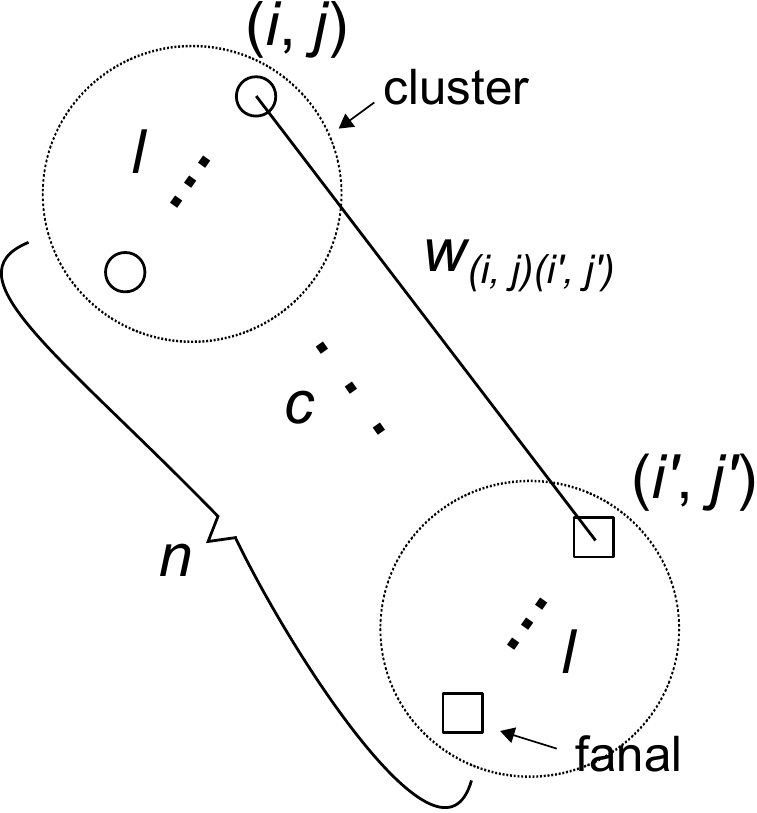}%0.45\textwidth
\caption{The network general structure and notation}
\label{Figure 1}
\end{figure}

The considered network relies on binary neurons and binary connections. The authors of \cite{GriBer20117} use the term \textit{fanal} instead of neuron to underline their biological inspiration. Figure~\ref{Figure 1} represents the general structure of the network and the notation. All the $n$ fanals are organized in $c$ disjoint groups called \textit{clusters}. Fanals belonging to specific clusters are represented with different shapes. Each cluster groups $l=n/c$ fanals, when all the clusters are of the same size. Note that the equal sizes of the clusters are not required for the network functioning. A node in the network is identified by its index $(i, j)$, where $i$ corresponds to the cluster number and $j$ to the number of the fanal inside the cluster. The connections are allowed only within fanals belonging to different clusters, i.e. the graph is multipartite. The connection between two fanals is denoted $w_{(i, j)(i', j')}$. Contrary to classical neural networks the connections do not possess different weights, the connection exists or not. A particular connection is established by convention if its weight equals 1, 0 otherwise.

A given cluster corresponds to a particular part of a message $m$ to store, whereas a given fanal inside the cluster represents a particular value for this part. Storing messages in the network is realized by connecting the fanals from all the clusters materializing different parts of the message to store. Such fully interconnected subgraph represents the message in the network and is called a \textit{clique}. For the picture clarity Figure~\ref{Figure 1} does not represent the full clique which is illustrated in the following section.

After storing a message, retrieval process is organized as follows. First, the known parts of the message are used to stimulate appropriate fanals. After the initial stimulation, message passing phase comes next. The activated fanals send unitary signals to other clusters through all of their connections. Then, each of the fanals calculates the sum of the signals it received. Within each cluster the fanal having the highest sum is chosen and its state becomes 1. The rest of the fanals inside the cluster present the state equal to 0. The rule according to which the active fanal inside the cluster is elected is called \textit{Winner Takes All} (WTA). The whole process may be iterative such that ambiguous clusters (those containing more than one active fanal) will hopefully be correctly retrieved. Thanks to the strong correlation brought by the connections, it is possible to retrieve the stored message based on partial information put into the network. Though it is correlation inside each stored message that is actually used to retrieve it using the network, the correlation between distinct messages can lead to dramatic loss in performance as explained in the following section.

\section{Non-uniformly distributed messages}
\label{sec:non-uniformly distributed messages}
\subsection{Density and error probability definition}

As previously stated storing messages in the network corresponds to creating subgraphs of interconnected fanals. When the number of stored messages increases, these subgraphs share an increasing number of connections. Consequently, distinguishing between messages is more difficult. The network density $d$ is defined as a ratio of established connections to all the possible ones. Therefore, the density is a parameter of first importance for the network performance. A density close to 1 corresponds to an overloaded network. In this case the network will not be able to retrieve stored messages correctly. For a network that stored $M$ uniformly distributed messages expected density $d$ is expressed by the following formula:

\begin{equation}
\label{density}
d = 1 - \left(1 - \frac{1}{l^2}\right)^M \approx \frac{M}{l^2} \text{ when } M \ll l^2.
\end{equation}
When clusters are not of the same size (i.e. the same number of fanals), $l^2$ is replaced with a product of $l_1$, $l_2$ of the clusters between which the density is calculated, where $l_1$, $l_2$ are the numbers of the fanals in these clusters. As stated before, the high density may lead to difficulties in distinguishing between messages. The probability of incorrectly retrieving a message with $c_e$ positions erased in a network constructed of $c$ clusters is given by:

\begin{equation}
\label{error probability}
P_e = 1 - \left(1 - d^{c-c_e}\right)^{(l - 1) c_e}.
\end{equation}
The probability of error increases with $d$, which is expressed by \eqref{density}. The equation \eqref{error probability} is valid for a single iteration. Note that iterations improve the ability of the network for retrieving messages correctly.

\subsection{Non-uniform distribution example}

\begin{figure}
\centering
\includegraphics[width=70mm]{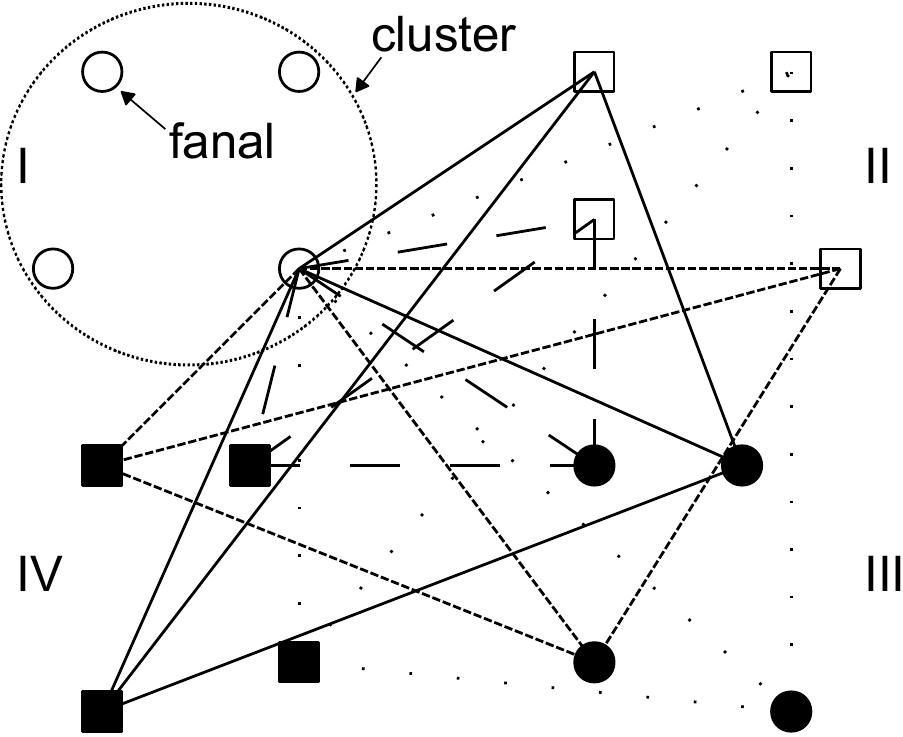}%0.45\textwidth
\caption{Network with non-uniformly distributed messages}
\label{Figure 2}
\end{figure}

Figure~\ref{Figure 2} represents an exemplary network. As before, fanals belonging to specific clusters are represented with different shapes. There are four fanals in each of the four clusters. Therefore, the length of the messages is four, each cluster corresponding to a given position, and on each position one of the four values is chosen. Cliques are formed by lines connecting fanals. Figure~\ref{Figure 2} depicts a situation where four messages are stored.

Let us suppose now that on a given position in a set of messages to store one of the values is much more frequent than the others (Figure~\ref{Figure 2}). Since this value corresponds to one particular fanal, this node will have much more connections than the rest in its cluster. One can observe that in all of the clusters except the top-left, each node has the same number of connections. This means that on the positions II, III, IV of the messages each of the four possible values occurs the same number of times. However, in the top-left cluster I only one of the fanals is always used, i.e. the value on the first position is constant. One may observe many connections going out from this node.

\begin{figure}
\centering
\includegraphics[width=90mm]{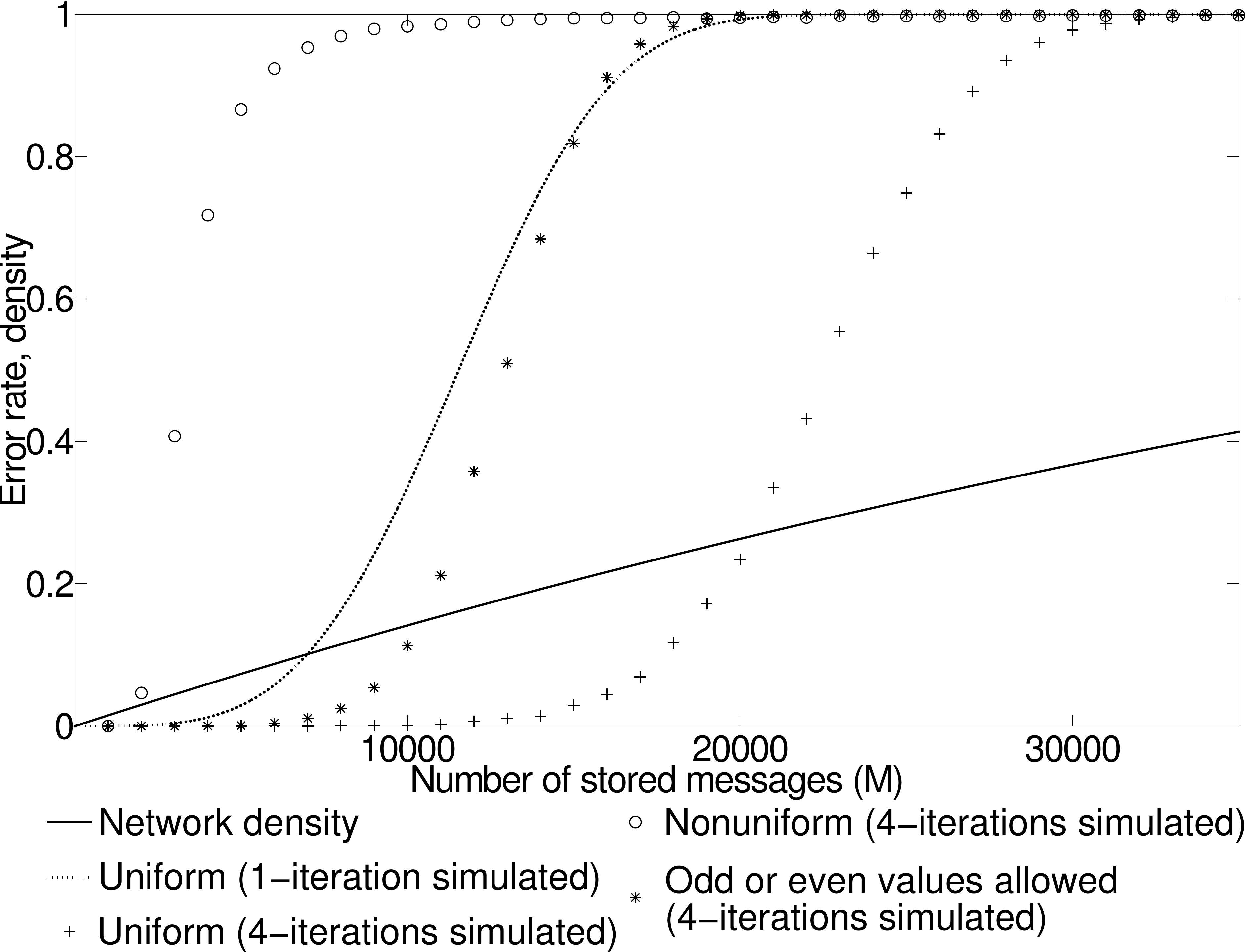}
\caption{Evolution of the error rate as a function of the number of stored messages. The network composed of eight clusters of size 256, four randomly erased positions}
\label{Figure 3}
\end{figure}

This simple example shows how the distribution of values present in data stored in the network maps to interconnection structure. Non-uniform distribution of dataset produces high density areas which may dramatically reduce the retrieval performance. Assume a network with $c=8$ and $l=256$. Figure~\ref{Figure 3} shows the evolution of the error retrieval rate when half the clusters are not provided with information. The simulation shows that the network of $n=2048$ fanals can store up to 15000 uniform messages of 64 bits each and retrieve them with a very high probability under strong erasures (error rate 0.029). However, when the messages are generated from Gaussian distribution, error rate exceeds the value for uniform case when only 2000 messages are stored (error rate 0.047). Figure~\ref{Figure 3} presents also a curve for a data with a specific correlation between the values within each message. Inside each message either odd or even values are only allowed. This means that if on the first position there is an odd value, one knows that all the other values are odd. For this dataset the network can store 8000 messages and retrieve them with the same error probability as in the uniform case. Figure~\ref{Figure 3} shows also the theoretical curve for a single iteration and the network density. Note the interest of the iterative character of the decoding process. The presented results clearly indicate the importance of stored data distribution.

For non-uniform data, equation \eqref{density} is not valid since it assumes uniformly distributed messages. It is worth mentioning an extreme example of non-uniformly distributed messages which is a set of constant values. The retrieval error rate in this case is 0. However, as these messages share the same connections they do not produce local high density areas.

The aim of this work is to enrich the analysis regarding retrieval rate with respect to $d$ performed in \cite{GriBer20117} with the analysis of the data distribution.

\section{Strategies to store non-uniform data}

In the following section several strategies to solve the problem of storing non-uniform data are described.

\subsection{Adding random clusters}

\begin{figure}
\centering
\includegraphics[width=63mm]{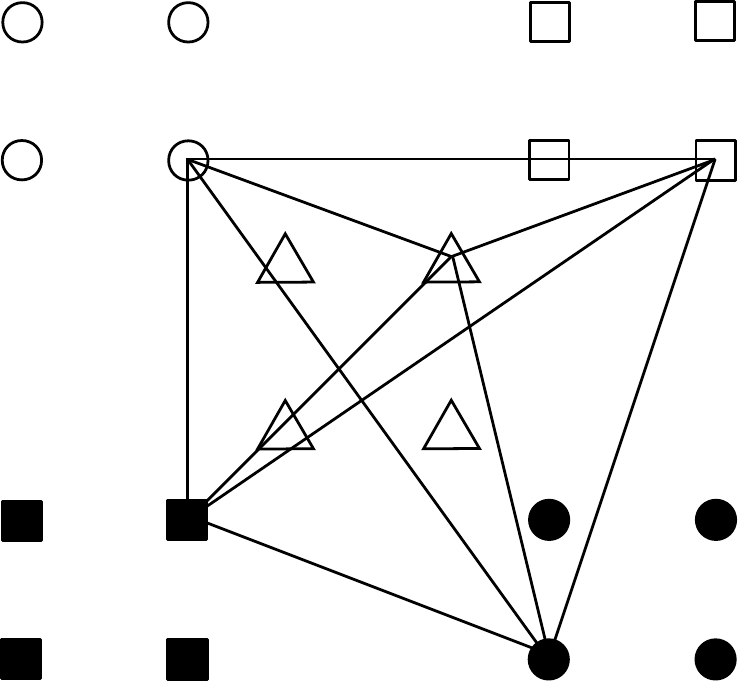}
\caption{Network with one random cluster (represented by triangles) added}
\label{Figure 4}
\end{figure}

The first of the strategies relies on adding to the existing network clusters filled with random values drawn from a uniform distribution. These random values stored in \textit{random clusters} play a role of so-called \textit{stamp}, providing the existing clique with additional information and supporting the message retrieval process. This way the influence of local high density areas caused by non-uniform data is neutralised. Figure~\ref{Figure 4} illustrates the network from Figure~\ref{Figure 2} with one random cluster added and one message stored.

Similar technique is introduced in \cite{Knoblauch:2010}. The authors propose adding to a feedforward neural network an additional intermediary layer filled with random patterns, in order to improve the performance of a single-layer model in case of non-random data. 

The following section presents how other techniques proposed in this work improve the performance compared to adding random clusters.

\subsection{Adding random bits}

Another category of the proposed strategies relies on adding random uniformly generated bits to the existing data rather than whole random values stored in separate clusters. As a consequence, the structure of the network remains the same, only the size of clusters is modified since the range of values is expanded by a number of random bits.

\begin{figure}
\centering
\includegraphics[width=90mm]{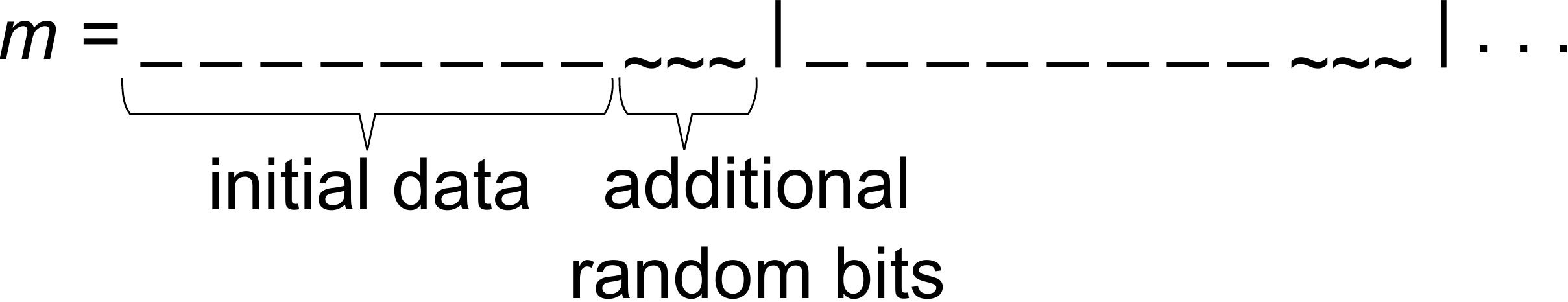}
\caption{Adding random bits. Random bits represented by {\raise.17ex\hbox{$\scriptstyle\mathtt{\sim}$}}, initial data bits represented by -}
\label{Figure 5}
\end{figure}

Figure~\ref{Figure 5} presents how random bits are added to each value forming a message. Each character in a message is coded on eight bits and expanded with three random bits represented by tilde.

\begin{figure}
\centering
\includegraphics[width=4cm]{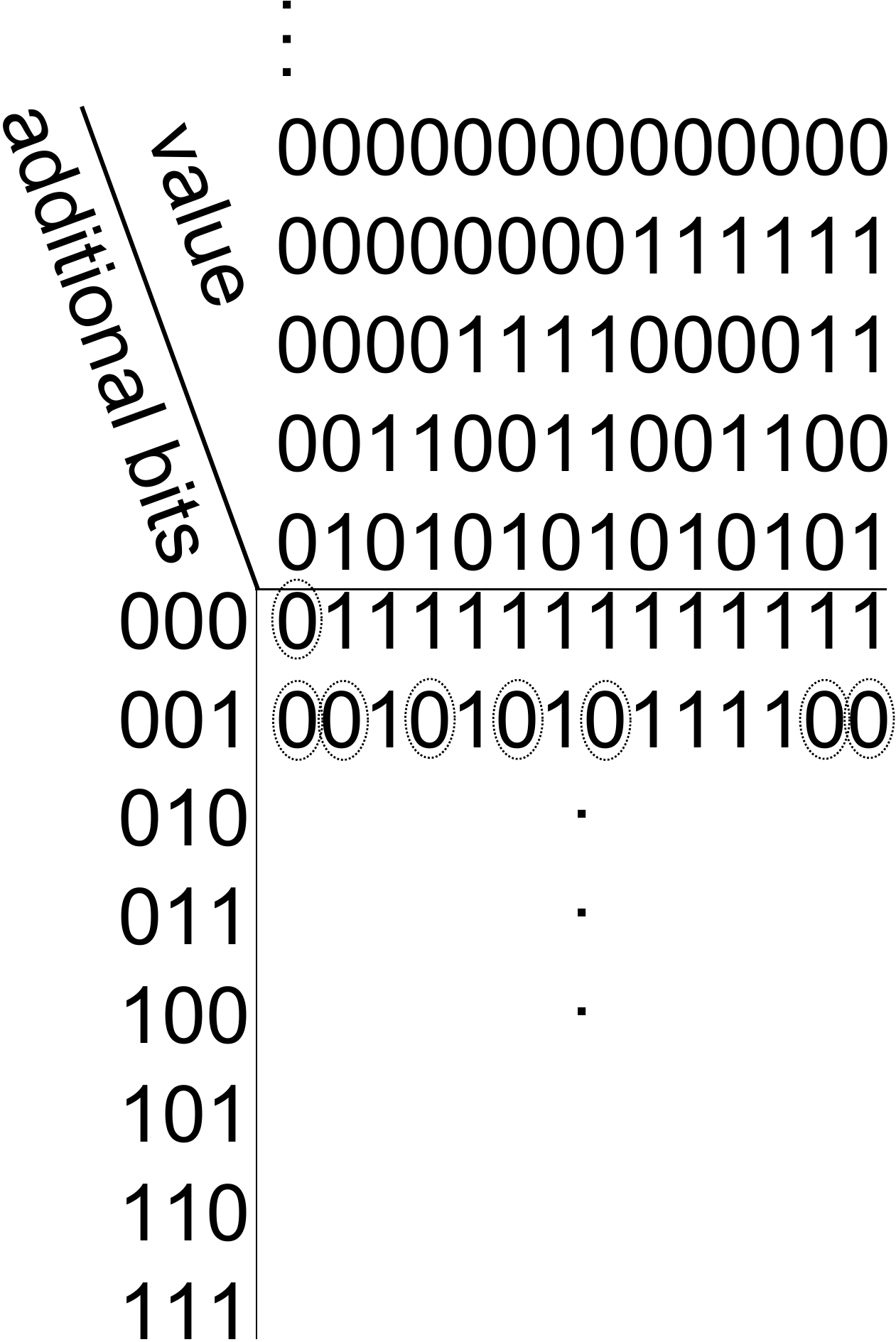}
\caption{Adding least used combination of bits}
\label{Figure 6}
\end{figure}

Besides adding bits simply drawn from uniform random distribution another approach to generate additional bits is proposed. To each value always the least used combination of bits (or one of the least used) is added. The rule is illustrated with Figure~\ref{Figure 6}. For each of the values coded on 8 bits a table is created where the number of occurrences of each additional bits combination is stored. For instance, for the first value consisting of only zeros either 000 or 001 can be chosen. However, for value 00000001 only 001 can be added since 000 is already used once. This procedure continues for all positions in all the messages.

\subsection{Using compression codes}

The last proposed strategy is to apply algebraic compression codes. In this work Huffman lossless compression coding  is proposed. This technique allows to minimize the average number of coding symbols per message \cite{huf52}. Note that for some datasets arithmetic coding may perform better than Huffman code, see \cite{Bookstein93} for comparison.

Huffman coding produces variable length codewords - the values that occur most frequently are coded on a small number of bits, whereas less frequent values occupy more space. Therefore, the sizes of frequent values that break the uniformity are minimized. The obtained free space is filled with random uniformly generated bits. Decoding is possible thanks to the prefix-free property, that is a set of bits representing a symbol is never a prefix of another codeword.

\section{Performance comparison}
\subsection{Evaluation of the proposed strategies}

In this section performance of all the introduced strategies is evaluated. The simulations are performed for the same network as in section~\ref{sec:non-uniformly distributed messages} and the same non-uniform distribution.

\begin{figure}
\centering
\includegraphics[width=90mm]{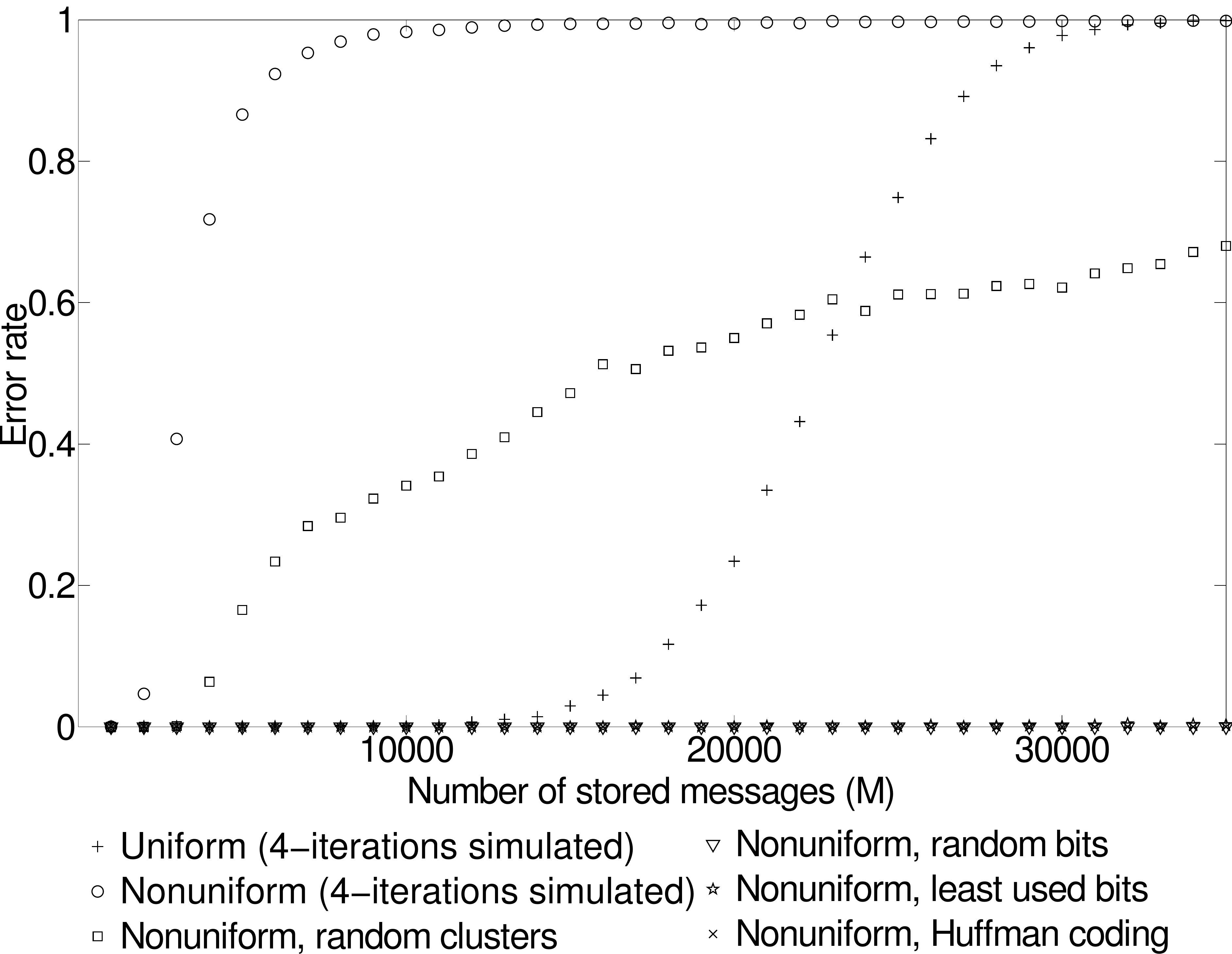}
\caption{Evolution of the error rate when using proposed strategies as a function of the number of stored messages compared to uniform and non-uniform cases. The material used for each strategy is comparable}
\label{Figure 7}
\end{figure}

Figure~\ref{Figure 7} shows the improvement in performance when using the proposed techniques. To prove that adding a certain material improves the network functioning seven random clusters of 5000 fanals each are added. Compared to non-uniform case without using any technique the improvement is clear. The curve for random clusters crosses the uniform case at a certain number of stored messages. This is because the number of messages the network can store and then retrieve becomes linear. The function is non-monotonic since the significant part of the network is filled with random values.

The rest of the techniques use the closest additional material compared to the random clusters strategy, considered as the number of possible connections. This is obtained for eight clusters of 4096 fanals which corresponds to four additional bits. These strategies offer much better performance using comparable material.

\begin{figure}
\centering
\includegraphics[width=90mm]{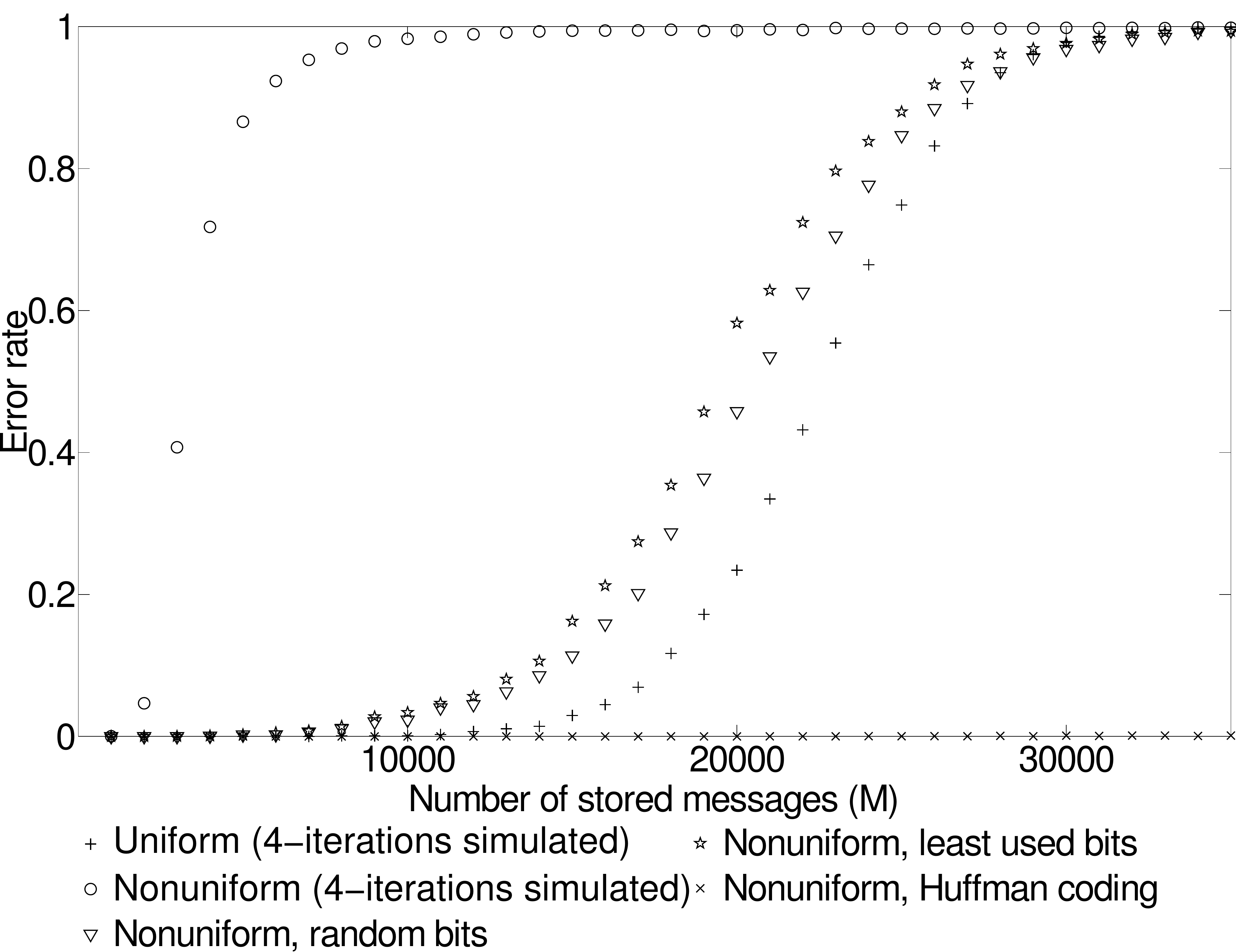}
\caption{Evolution of the error rate when using proposed strategies as a function of the number of stored messages compared to uniform and non-uniform cases. The material used for each strategy is equal}
\label{Figure 8}
\end{figure}

Figure~\ref{Figure 8} presents results when using minimum material to approach the performance close to the uniform case. When random bits and least used combination strategies are applied two additional bits are enough to get close to uniform performance. This means that the size $l$ of the clusters equals 1024 and the material used accounts for 4.9\% of the material used for random clusters strategy. When three bits are added the performance gets already much better than for uniform messages. The curve for Huffman coding technique also uses two additional bits. In this case the gain in performance is significant. For the used data distribution the length of the messages sometimes exceeds 72 bits and consequently, adding less than two bits is not possible. However, for different data sets minimizing the used material could still be feasible.

\subsection{Application example}

In \cite{LarrasBoguslawski:2013} authors propose a power management technique for Multiprocessor System-on-Chip (MPSoC) based on the theory of \cite{GriBer20117}. Depending on the current working conditions and the data stored in the network the frequency for each Voltage/Frequency Island (VFI) is assigned. This allows dynamically adapting the MPSoC power distribution to the current tasks and therefore reduce the energy consumption. In this approach a subset of network clusters is associated with the working conditions and the remaining clusters provide the information about the frequency values. It is however unlikely that the values of working conditions and frequencies are uniformly distributed. Assume for example a demanding application where in most cases high frequency values are chosen. This implies storing more cliques connected to a subgroup of fanals and produces local high density areas. By using the strategies proposed in the present work one can neutralize the influence of non-uniform data distribution and improve the performance of the technique proposed in \cite{LarrasBoguslawski:2013}.

\section{Conclusion}

In this work the performance of the network \cite{GriBer20117} in case of non-uniformly distributed messages is evaluated. The analyses showed clearly the importance of adapting the network to this kind of data. Several strategies to avoid performance degradation in this case are proposed and evaluated. Later, an application example is given. In order to apply this kind of memory in practical applications adapting the network to non-uniform messages is indispensable.

The strategy that uses Huffman coding offers great performance improvements. However, it needs constructing the dictionaries in order to code and decode messages. Future work may include exploring some self-adapting techniques built in the network architecture. For instance, one can envision limiting the density on fanal level and when a specified value is reached using a new fanal instead. Another possibility to be considered is adapting the size of cliques with respect to data distribution.



%% References


\bibliographystyle{elsarticle-num}
\bibliography{references}

\begin{thebibliography}{10}
\expandafter\ifx\csname url\endcsname\relax
  \def\url#1{\texttt{#1}}\fi
\expandafter\ifx\csname urlprefix\endcsname\relax\def\urlprefix{URL }\fi
\expandafter\ifx\csname href\endcsname\relax
  \def\href#1#2{#2} \def\path#1{#1}\fi

\bibitem{Lin:1976}
C.~S. Lin, D.~C.~P. Smith, J.~M. Smith, The design of a rotating associative
  memory for relational database applications, ACM Trans. Database Syst. 1~(1)
  (1976) 53--65.

\bibitem{Papadogiannakis:2010}
A.~Papadogiannakis, M.~Polychronakis, E.~P. Markatos, Improving the accuracy of
  network intrusion detection systems under load using selective packet
  discarding, in: Proceedings of the Third European Workshop on System
  Security, EUROSEC '10, New York, NY, USA, 2010, pp. 15--21.

\bibitem{Jouppi:1990}
N.~P. Jouppi, Improving direct-mapped cache performance by the addition of a
  small fully-associative cache and prefetch buffers, in: Proceedings of the
  17th annual international symposium on Computer Architecture, ISCA '90, New
  York, NY, USA, 1990, pp. 364--373.

\bibitem{Huang:2001}
N.~H. W. C.~C. Lou, J.~Chen, Design of multi-field {IP}v6 packet classifiers
  using ternary cams, in: Proc. IEEE GLOBECOM, Vol.~3, 2001, pp. 1877--1881.

\bibitem{GriRab20139}
V.~Gripon, M.~Rabbat, Maximum likelihood associative memories, in: Proceedings
  of Information Theory Workshop, 2013, submitted to.

\bibitem{pagiamtzis-jssc:2006}
K.~Pagiamtzis, A.~Sheikholeslami, Content-addressable memory ({CAM}) circuits
  and architectures: A tutorial and survey, IEEE Journal of Solid-State
  Circuits 41~(3) (2006) 712--727.

\bibitem{JarollahiGripon:2013}
H.~Jarollahi, V.~Gripon, N.~Onizawa, W.~J. Gross, A low-power
  content-addressable-memory based on clustered-sparse-networks, in:
  Proceedings of 24th IEEE International Conference on Application-specific
  Systems, Architectures and Processors (ASAP 2013), Washington, DC, USA, 2013.

\bibitem{Agrawal06tcam}
B.~Agrawal, T.~Sherwood, Modeling {TCAM} power for next generation network
  devices, in: Proc. of IEEE International Symposium on Performance Analysis of
  Systems and Software (ISPASS), Austin, TX, 2006, pp. 120--129.

\bibitem{HopfTank:82}
J.~J. Hopfield, Neural networks and physical systems with emergent collective
  computational abilities, Proc. Nat. Acad. Sci 79~(8) (1982) 2554--2558.

\bibitem{willshaw:nonholographic}
D.~J. Willshaw, O.~P. Buneman, L.~H.~C. Higgins, Non-holographic associative
  memory, Nature 222 (1969) 960--962.

\bibitem{Palm:2013}
G.~Palm, Neural associative memories and sparse coding, Neural Netw. 37 (2013)
  165--171.

\bibitem{GriBer20117}
V.~Gripon, C.~Berrou, Sparse neural networks with large learning diversity,
  IEEE Transactions on Neural Networks 22~(7) (2011) 1087--1096.

\bibitem{AcklHint:85}
D.~H. Ackley, G.~E. Hinton, T.~J. Stejnowski, A learning algorithm for
  {B}oltzmann machines, Cognit. Sci. 9~(1) (1985) 147--169.

\bibitem{Knoblauch:2010}
A.~Knoblauch, G.~Palm, F.~T. Sommer, Memory capacities for synaptic and
  structural plasticity, Neural Comput. 22~(2) (2010) 289--341.

\bibitem{huf52}
D.~A. Huffman, A method for the construction of minimum-redundancy codes,
  Proceedings of the Institute of Radio Engineers 40~(9) (1952) 1098--1101.

\bibitem{Bookstein93}
A.~Bookstein, S.~Klein, Is {H}uffman coding dead?, Computing 50 (1993)
  279--296.

\bibitem{LarrasBoguslawski:2013}
B.~Larras, B.~Boguslawski, C.~Lahuec, M.~Arzel, F.~Seguin, F.~Heitzmann, Analog
  encoded neural network for power management in {MPSoC}, in: Proceedings of
  the 11th International IEEE New Circuits and Systems Conference, NEWCAS '13,
  2013.

\end{thebibliography}






% \end{thebibliography}

\parpic{\includegraphics[width=1in,clip,keepaspectratio]{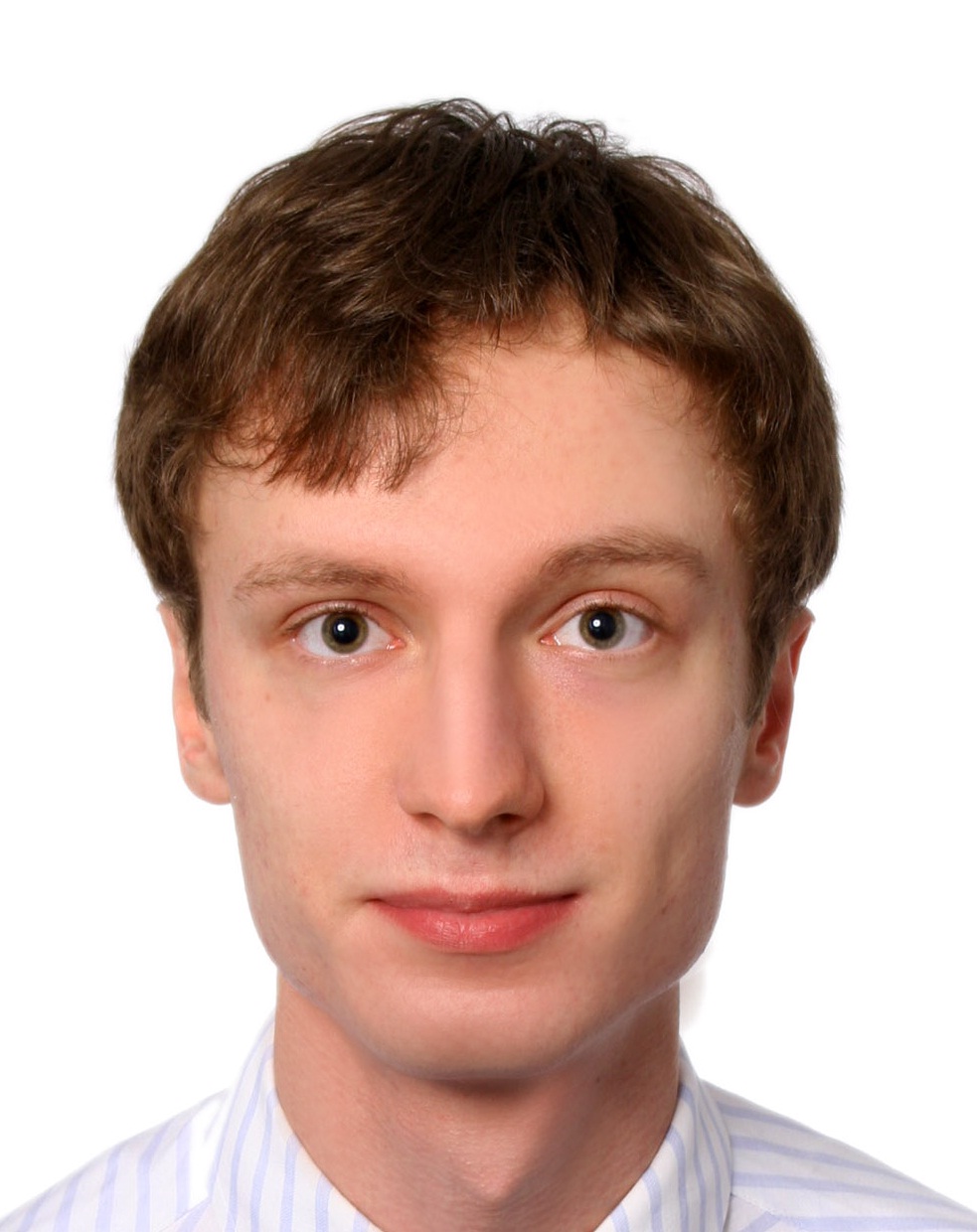}}
\noindent {\bf Bartosz Boguslawski} obtained his B.S. and M.S degrees in Electronics and Telecommunications from AGH University of Science and Technology, Cracow, Poland, in 2011 and 2012 respectively. He was an exchange student at Ghent University, Belgium where he prepared his Master thesis in the field of inverse problems for heat transfer in microelectronics. He is currently a Ph.D. candidate at CEA-Leti, Grenoble and TELECOM Bretagne, Brest, France, directed by Prof. Claude Berrou. His current research concerns power management, multiprocessor system-on-chip, and neural networks.

\parpic{\includegraphics[width=1in,clip,keepaspectratio]{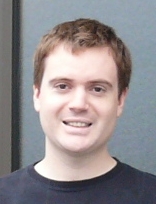}}
\noindent {\bf Vincent Gripon} obtained his Ph.D. in 2011 with TELECOM Bretagne (Institut Mines-T\'{e}l\'{e}com), Brest, France, under the supervision of Professor Claude Berrou. He is now a postdoc and he specializes in computer and information sciences. His work mainly focuses on connecting information theory and error correcting codes with neural networks. His main contribution is a new family of sparse associative memories based on distributed codes that provide almost optimal efficiency. He is also the creator and organizer of a programming contest named TaupIC, which targets French top undergraduate students.

\parpic{\includegraphics[width=1in,clip,keepaspectratio]{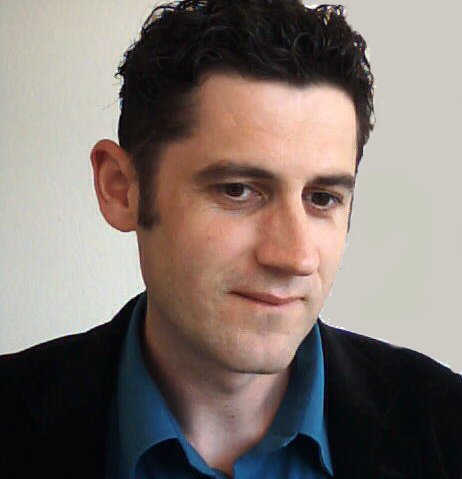}}
\noindent {\bf Fabrice Seguin} was born in Talence, FRANCE in 1973. He received the Ph.D. degree from the Universite Bordeaux 1, France, in 2001. His doctoral research concerned the current mode design of high-speed current-conveyors and applications in RF circuits. In 2002, he joined the Electronic Engineering Department of TELECOM Bretagne, Brest, France, as a full-time lecturer. He is currently involved with design issues of analogue channel decoders and related topics, energy harvesting, and neural networks.

\parpic{\includegraphics[width=1in,clip,keepaspectratio]{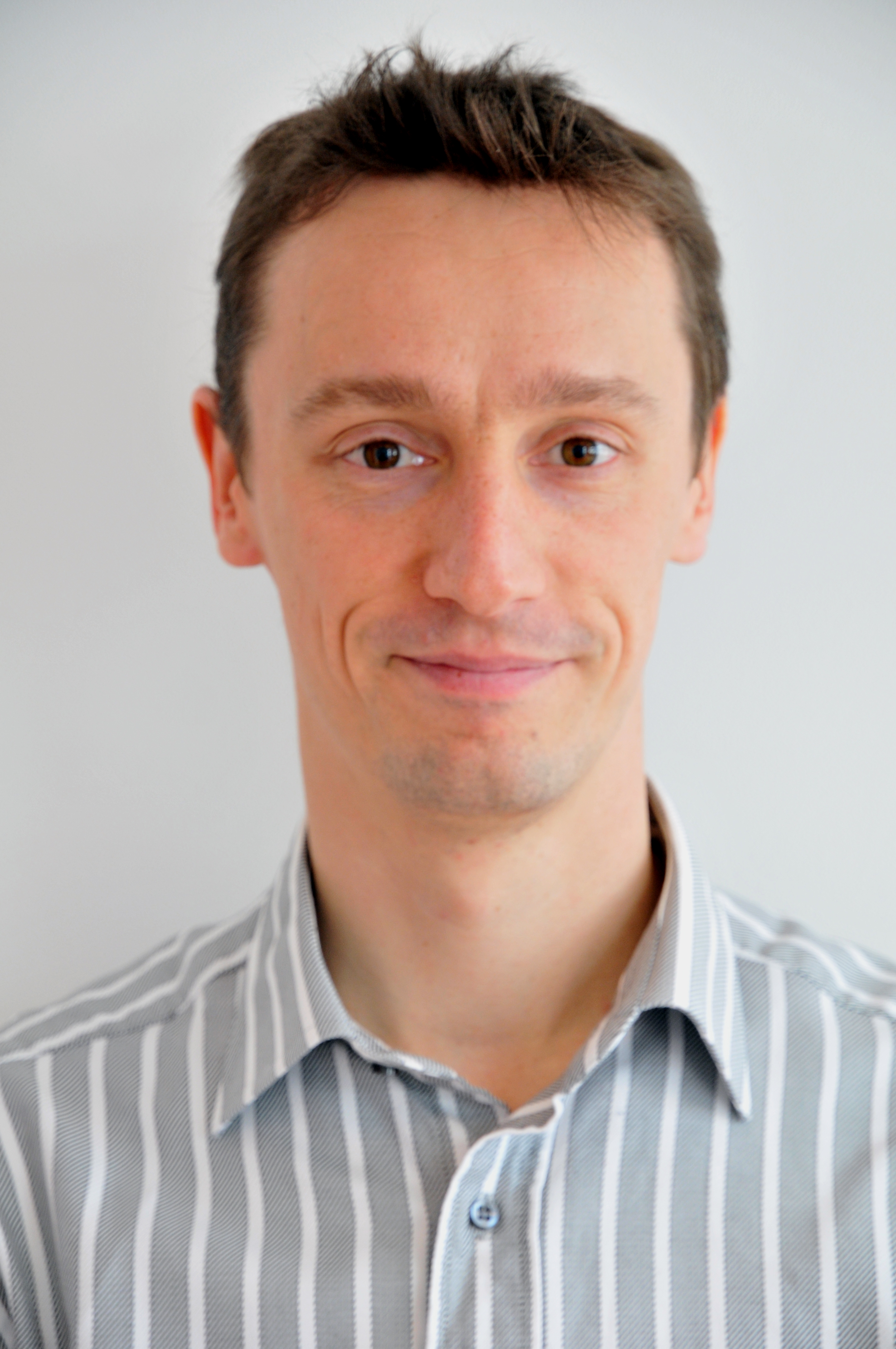}}
\noindent {\bf Fr\'{e}d\'{e}ric Heitzmann} received the M.S. degree from the Ecole polytechnique, Palaiseau, France, and the M.S. degree from the Telecom Paritech, Paris, France, specialized in computer science and networks.
His research interests include hardware-software co-design, and compilers for heterogeneous and homogenous MPSoC.

\clearpage
\listoffigures

\end{document}